\documentclass[runningheads]{llncs}
\usepackage{graphicx}
\usepackage{amsmath}
\usepackage{amssymb}
\usepackage{algorithm}
\usepackage{algpseudocode}
\usepackage[hidelinks]{hyperref}
\usepackage{subcaption}
\begin{document}
\title{Inverse Reinforcement Learning With Constraint Recovery}%\thanks{Supported by organization x.}}
%
%\titlerunning{Abbreviated paper title}
% If the paper title is too long for the running head, you can set
% an abbreviated paper title here
%
\author{Nirjhar Das%\orcidID{0000-1111-2222-3333} 
\and
Arpan Chattopadhyay}%\inst{1}}%\orcidID{1111-2222-3333-4444} \and
%Third Author\inst{3}\orcidID{2222--3333-4444-5555}}

\authorrunning{N. Das and A. Chattopadhyay}
% First names are abbreviated in the running head.
% If there are more than two authors, 'et al.' is used.
%
\institute{Indian Institute of Technology Delhi, New Delhi, India \\
\email{\{Nirjhar.Das.ee319, arpanc\}@ee.iitd.ac.in}}

\maketitle              % typeset the header of the contribution
\begin{abstract}
In this work, we propose a novel inverse reinforcement learning (IRL) algorithm for constrained Markov decision process (CMDP) problems. In standard IRL problems, the inverse learner or agent seeks to recover the reward function of the MDP, given a set of trajectory demonstrations for the optimal policy. In this work, we seek to infer not only the reward functions of the CMDP, but also the constraints. Using  the principle of maximum entropy, we  show that the IRL with constraint recovery (IRL-CR) problem can be cast as a constrained non-convex optimization problem. We reduce it to an alternating constrained optimization problem whose sub-problems are convex. We use exponentiated gradient descent algorithm to solve it. Finally, we demonstrate the efficacy of our algorithm for the grid world environment.

%\keywords{Reinforcement Learning  \and Inverse Reinforcement Learning \and Constrained Markov Decision Process \and Constrained Optimization.}
\end{abstract}

\section{Introduction}
%\subsection{A Subsection Sample}
%Please note that the first paragraph of a section or subsection is
%not indented. The first paragraph that follows a table, figure,
%equation etc. does not need an indent, either.

%Subsequent paragraphs, however, are indented.

%The goal of Artificial Intelligence is to produce agents that can learn from their surroundings and make their own decisions. Reinforcement Learning is a promising direction in which the goal of the agent is to learn an optimal policy by interacting with the environment. Reinforcement Learning is often applied in a standard Markov Decision Process setting. A Markov Decision Process (MDP) comprises of a state space and an action space. The agent gets a reward when it takes a particular action in a particular state. The rewards are usually stochastic in nature. Thereafter, the agent moves from the current state to the next state based on a state-transition probability distribution. The process can go on for finite as well as an infinite number of steps. The goal of the agent is to learn an optimal policy that helps the agent exploit the environment by collecting maximum return. Reinforcement Learning is of great interest to researchers and several algorithms have been proposed to learn an optimal policy in various settings. Some of the popular algorithms are Policy Gradient\cite{pg}, Actor-Critic Methods\cite{ac}, Trust Region Policy Optimization\cite{trpo}, Deep Q-Networks\cite{dqn}, Dyna\cite{dyna} and AlphaZero\cite{alphazero}.

Reinforcement Learning (RL, \cite{rl-survey,sutton-barto}) is a popular branch of artificial intelligence where an agent makes sequential decisions under uncertainty. Every time, the agent observes the \textit{state} of the environment, chooses an \textit{action} based on the current state, obtains a \textit{reward} and transits  to a random next state based on a probability distribution depending on the current state-action pair. The agent's goal is to choose the actions to maximize the expected sum of rewards over a time horizon; this gives rise to the problem of finding the optimal \textit{policy} that, given the state at any time, prescribes an action to the agent. When the transition probability or reward function are unknown to the agent, RL is used  to learn the optimal policy. RL has massive  applications in various domains like autonomous driving, drug discovery, resource allocation in communication systems, robot learning, path planning, and   large language models.

Inverse Reinforcement Learning (IRL) comes under the broad paradigm of learning from demonstrations~\cite{LFD,survey-lfd} where the goal is to learn behaviour from given expert demonstrations. Here, an inverse agent seeks to  learn the reward function (and hence the objective) of an RL problem encountered by a forward agent. Typically, the inverse agent is provided with a set of trajectories of the RL problem of the forward agent.   IRL is applied in several applications, such as  autonomous driving and robotic manipulations. 
IRL is inherently ill-posed as several reward functions may conform with the same set of demonstrations. However, there have been several works on IRL in the literature, such as  \cite{irl-ng} where IRL is posed as a linear program,  \cite{ratliff-06} where IRL is posed as a structured maximum margin classification, and importantly \cite{maxentirl} that proposes the new paradigm of maximum entropy IRL   (MaxEnt IRL) that assumes that the demonstrated     trajectories are sampled from a Boltzmann  distribution with the negative sum reward of a sample trajectory as its potential function. In fact, several works \cite{wulfmeier-15,finn-16,ho-ermon-16} have used this maximum entropy principle to demonstrate strong performance on large-scale data sets.

In this work, we consider IRL where the forward agent encounters a constrained Markov decision process (CMDP \cite{altman}) problem, and the inverse agent seeks to recover both the reward and the constraint functions from a set of demonstrations.  Many real-life tasks involve solving a CMDP. For example, while pouring water from a glass jar into a glass bottle, one must satisfy the constraint that neither glass vessel breaks. Similarly, path planning of an autonomous vehicle should ensure that the manoeuvring decision does not lead to an accident with neighbouring vehicles.  Obviously, recovering the constraint function along with the reward function enables the IRL agent to obtain a  realistic picture of the objectives of the forward agent. 

Recent works on  IRL for CMDP can be broadly categorized into two  classes: (a) algorithms that recover rewards or policies given the knowledge of constraints and (b) algorithms that recover constraints given the reward function. The papers \cite{inverse-kkt,kalweit-20,x-men-22} belong to the first group, while \cite{scobee2020,malik2021,chou-20,chou-21,park-20,papadimitriou-21,gaurav-22} belong to the second group.  In \cite{x-men-22}, the reward function is learnt when known multi-state combinatorial constraints restrict the distribution of trajectories. On the other hand, \cite{inverse-kkt} considers learning the reward function when demonstrations are generated by the forward agent by using  optimal control for the constrained problem. In \cite{kalweit-20}, the authors propose an inverse Q-learning algorithm that updates the Q-estimates using the knowledge of the constraints. A notable work in the second class is \cite{scobee2020}, where the goal is to recover a maximum likelihood constraint set, which, when appended to the base MDP with known reward, conforms with the observed trajectories. In \cite{malik2021}, the authors extend \cite{scobee2020} to a parameterized policy setting with the objective to recover the forward agent's policy. In \cite{gaurav-22}, the authors propose an algorithm to learn soft constraints that are bounded in expectation, from demonstrations in stochastic setting with continuous states and actions and known rewards. Moreover, \cite{chou-21,park-20,papadimitriou-21} take Bayesian approaches to constraint learning when rewards are known.

In this paper, we consider the IRL problem where the inverse agent seeks to recover both the reward and constraint functions of the forward agent, given that the trajectory demonstrations  have been obtained by executing an optimal policy for the forward agent's CMDP.  We call this problem IRL with constraint recovery (IRL-CR). This problem is very important for scenarios that involve  CMDPs, such as autonomous driving, healthcare and other safety-critical applications. 
To the best of our knowledge, this problem has not been solved before, and the authors of \cite{gaurav-22} claim that this problem is difficult to solve due to its ill-posed nature.   However, we formulate a constrained optimization problem using the principle of maximum entropy and maximum likelihood estimation to tackle this challenge and  provide  a simple and efficient algorithm that achieves strong numerical performance. The main contributions of our work are:
\begin{enumerate}
    \item We formulate a  novel constrained optimization problem  to recover the reward and the constraint simultaneously from demonstrations. Assuming a linear function approximation  of the reward and constraint functions and by using the maximum entropy principle, we  derive a Boltzmann distribution over trajectories parametrized by both the reward and constraint and use it to derive a non-convex constrained optimization problem. 
    \item We reduce the constrained, non-convex problem to an alternating constrained optimization problem whose sub-problems are convex.  We use Exponentiated Gradient Descent to derive a simple and efficient algorithm for its solution. 
    \item We   demonstrate strong empirical performance  in a grid world environment.
\end{enumerate}

\section{Preliminaries}
A CMDP is characterized by a septuple $(\mathcal{S}, \mathcal{A}, r, c, \alpha, p, \gamma)$, where $\mathcal{S}$ is the state space, $\mathcal{A}$ is the action space, $r$ is the reward function, $c$ is the constraint function, $p$ is the state-transition probability set and $\gamma \in (0,1)$ is a discount factor. When an agent in state $s\in \mathcal{S}$ takes action $a\in \mathcal{A}$, then the agent receives a reward $r(s,a)$, incurs a penalty $c(s,a)$, and transits to the next state according to the probability distribution $p(\cdot|s,a).$  A \textit{stationary, randomized policy} is a mapping $\pi:\mathcal{S}\rightarrow \triangle(\mathcal{A})$, where $\triangle(\mathcal{A})$ is the probability simplex over $\mathcal{A}$. Obviously,  $\pi(a|s)$, denotes the probability of taking action $a$ when the state is    $s$. The \textit{value functions} associated with the reward or the constraint is defined as:
\begin{equation}
    V_x^\pi(s)=\mathbb{E}_{\pi}\Big[\sum_{t=1}^T \gamma^{t-1}x(s_t, a_t)\Big|s_1=s\Big], \,\,\, x \in \{r,c\}
\end{equation}
where $\mathbb{E}_{\pi}$ denotes the expectation under policy $\pi$. The optimal policy $\pi^*$ solves:
\begin{equation}
\label{opt-pol}
  \begin{aligned}
   \max_\pi\quad \sum_{s\in\mathcal{S}}p_0(s)V_r^\pi(s) \quad\quad
    \text{s.t.} \quad \sum_{s\in\mathcal{S}}p_0(s)V_c^\pi(s) \leq \alpha
\end{aligned}  
\end{equation}
where $p_0(\cdot)$ is the {\em known} initial state distribution. 
In this paper, we assume  $\alpha > 0$ and  $c(s,a) \geq 0\ \forall\ s\in \mathcal{S},\ a\in \mathcal{A}$. Also, we consider only one constraint, though an extension of our algorithm to  more than one constraint is straightforward. 

We denote by $\tau=\{s_1, a_1, s_2, a_2,\cdots,s_T , a_T\}$ a generic trajectory of the RL agent under policy $\pi^*$. The goal in the IRL-CR problem for an inverse agent is  to learn both the reward $r(\cdot,\cdot)$ and the constraint $c(\cdot,\cdot)$ functions  from a set of demonstrated trajectories $D=\{\tau_1, \dots,\tau_m\}$ under $\pi^*$. However, we assume that the inverse agent knows the tuple $(\mathcal{S}, \mathcal{A}, p, \gamma)$.

\section{Formulation}
The maximum likelihood (ML) estimates of the unknown reward and cost functions are: $(r^*, c^*, \alpha^*) = \text{arg}\max_{r,c,\alpha} \prod_{\tau\in D}p(\tau|r, c, \alpha)$, assuming that the demonstrated trajectories were generated by the RL agent under $\pi^*$. 

First, we normalize the constraint function $c(\cdot,\cdot)$ to obtain a new constraint $c'(\cdot,\cdot)=c(\cdot,\cdot)/\alpha$,  which eliminates the need to estimate  $\alpha$. In \eqref{opt-pol}, we can see that the optimal policy is unchanged if this normalization is performed. Hence, our objective is to solve (subject to the constraint in~\eqref{opt-pol}):
\begin{equation}
    \label{goal-rc}
    r^*, c^* = \text{arg}\max_{r,c'} \prod_{\tau\in D}p(\tau|r, c', 1)
\end{equation}
 Now, we make some assumptions about the reward and constraint  of the CMDP. 
\begin{enumerate}
    \item We assume that the reward and constraint are independent of the actions, that is: $r(s,a)=r(s)\forall\ s\in\mathcal{S},\ a\in\mathcal{A}$ and similarly for $c(\cdot,\cdot)$. This is a standard assumption in several prior works~\cite{maxentirl,wulfmeier-15,abeel-ng04} in IRL, and it reduces computational complexity. However, the method developed here can be easily generalized to cases where this assumption does not hold since the actions will still be observable to the inverse agent in those cases.
    \item We assume that the reward and cost functions are linear with respect to the features of the states denoted by $\Phi(\cdot)$; that is,  $r(s)=\mathbf{w}_r^\intercal \Phi_r(s),\ c(s)=\mathbf{w}_c^\intercal \Phi_c(s)$ 
    for some unknown parameters $\mathbf{w}_r\in\mathbb{R}^{d_r}$ and $\mathbf{w}_c\in\mathbb{R}^{d_c}$, where $d_r < |\mathcal{S}|$. Henceforth, with a slight abuse of notation, we will represent   $p(\tau|\mathbf{w}_r, \mathbf{w}_c)$ by $p(\tau)$, and hence,  $p(\tau|r,c,1)=p(\tau|\mathbf{w}_r,\mathbf{w}_c)$. This is a reasonable assumption since the inverse agent may be able to identify a set of meaningful feature vectors in realistic settings.
\end{enumerate}

Now, for $x \in \{r,c\}$, we define the quantities \textit{empirical feature expectation} (EFE) and \textit{policy feature expectation} (PFE) as follows:
\begin{equation}
\label{efe-pfe}
    \begin{aligned}
        \text{EFE:}\quad \tilde{\Phi}_x = \frac{1}{|D|}\sum_{\tau\in D} \Phi_x(\tau),\quad\quad
        \text{PFE:}\quad \hat{\Phi}_x = \sum_{\text{all}\ \tau} p(\tau)\Phi_x(\tau)
    \end{aligned}
\end{equation}
where $\Phi_x(\tau)=\sum_{s_t\in\tau}\gamma^{t-1}\Phi_x(s_t)$. In order to compute  the \textit{PFE} for any policy $\pi$, we use the \textit{state visitation frequency} defined as $\rho_\pi(s) = \sum_{t=1}^\infty \gamma^{t-1} \mathbb{P}_{\pi}(s_t=s)$. For a finite time horizon $T$, it can be approximated using the recursive equations:
\begin{equation}
    \label{svf-recur}
    \begin{aligned}
        d_1(s) &= p_0(s)\quad \forall s\in \mathcal{S}\\
        d_{t+1}(s') &= \sum_{s \in \mathcal{S}} \sum_{a \in \mathcal{A}}\gamma \cdot d_t(s) \cdot \pi(a|s) \cdot p(s'|s, a) \quad\forall s'\in \mathcal{S},\ \forall t\geq 1
    \end{aligned}
\end{equation}
With the help of the state visitation frequency, for $x\in\{r,c\}$, since $\rho_\pi(s) = \sum_{t=1}^\infty d_t(s)$., we can write (finite time approximation of) PFE as
\begin{equation}
\label{pfe-svf}
    \hat{\Phi}_x \approx \sum_{t=1}^T \sum_{s\in S} d_t(s)\Phi_x(s)
\end{equation}

\subsection{Obtaining the Boltzmann distribution}
In \cite{abeel-ng04}, the authors show that it is necessary and sufficient to match reward EFE and PFE to guarantee the same performance as the optimal policy under assumption (2), for the IRL problem without constraint recovery. We extend this to IRL-CR and ensure that the EFE and PFE for constraints also match. In formulating the optimization problem, we introduce a new constraint that  $p(\tau|\mathbf{w}_r,\mathbf{w}_r)$ satisfies the constraint in \eqref{opt-pol}. Next, we will use the principle of maximum entropy to determine the distribution $p(\tau|\mathbf{w}_r,\mathbf{w}_r)$ out of all possible distributions that satisfy these constraints. The problem can be formulated as:
\begin{equation}
\label{maxent-formulation}
    \begin{aligned}
        \min_p \sum_{\text{all}\ \tau} p(\tau)\log(p(\tau)) &
        \quad \text{s.t.}\quad \sum_{\text{all}\ \tau}p(\tau)\Phi_r(\tau) = \Tilde{\Phi}_r, \quad \sum_{\text{all}\ \tau}p(\tau)\Phi_c(\tau) = \Tilde{\Phi}_c\\
         &\quad \sum_{\text{all}\ \tau}p(\tau)\mathbf{w}_c^\intercal\Phi_c(\tau) \leq 1, \quad p(\tau) \in \triangle(\tau)
    \end{aligned}
\end{equation}
where $\triangle(\tau)$ is the probability simplex over all trajectories up to time $T$. This is a convex optimization problem in $p(\cdot)$. The Lagrangian of  $\mathcal{R}(p, \theta, \beta, \nu, \zeta, \delta)$ is:
\begin{equation}
\label{lagrangian}
    \begin{aligned}
        & \sum_{\text{all}\ \tau} p(\tau)\log(p(\tau)) + \theta^\intercal\Big(\sum_{\text{all}\ \tau}p(\tau)\Phi_r(\tau) - \Tilde{\Phi}_r\Big) + \beta^\intercal\Big(\sum_{\text{all}\ \tau}p(\tau)\Phi_c(\tau) - \Tilde{\Phi}_c\Big) \\ & \quad + \nu\Big(\sum_{\text{all}\ \tau}p(\tau)\mathbf{w}_c^\intercal\Phi_c(\tau) - 1\Big) + \zeta\Big(\sum_{\text{all}\ \tau}p(\tau)-1\Big) - \Big(\sum_{\text{all}\ \tau}\delta(\tau)p(\tau)\Big)
    \end{aligned}
\end{equation}
where $\theta, \beta, \nu, \zeta, \delta$ are Lagrange multipliers of appropriate dimensions. Also, note that $\nu\geq0$ and $\delta(\tau)\geq0\ \forall\tau$. From the KKT condition, we have $\frac{\partial\mathcal{R}}{\partial p(\tau)} = 0 \forall \tau$
% \begin{equation}
% \label{kkt-basic}
%     \begin{aligned}
%         \frac{\partial\mathcal{R}}{\partial p(\tau)} &= 0
%     \end{aligned}
% \end{equation}
under the optimal   $(p^*, \theta^*, \beta^*, \nu^*, \zeta^*, \delta^*)$. Hence, we obtain:
% \begin{equation*}
%     \begin{aligned}
%         1 + \log(p^*(\tau)) + \theta^{*T}\Phi_r(\tau) + \beta^{*T}\Phi(\tau) + \nu^* \mathbf{w}_c^\intercal\Phi_c(\tau) + \zeta^* - \delta^*(\tau) = 0
%     \end{aligned}
% \end{equation*}
% Solving for $p^*(\tau)$, we get
\begin{equation}
    \label{kkt-sol}
    \begin{aligned}
        p^*(\tau) = \exp(-1-\zeta^*+\delta^*(\tau)) \times \exp\big(-\theta^{*\intercal}\Phi_r(\tau) - [\beta^* + \nu^*\mathbf{w}_c]^\intercal\cdot \Phi_c(\tau)\big)
    \end{aligned}
\end{equation}

Now, we can interpret $-\theta^*$ as $\mathbf{w}_r$ since the higher the value of the term $-\theta^{*\intercal}\Phi_r(\tau)$, the higher will be the probability of that trajectory, thereby matching with the intuition that higher reward trajectories are more probable. Similarly, we can choose  $\beta^*=\mathbf{w}_c$. We make a further simplifying assumption that $\delta^*(\tau) = 0 \,\, \forall \tau$ since it can be assumed that $p^*(\tau) > 0\,\, \forall\tau$ and hence follows from complementary slackness condition. Thus, the expression \eqref{kkt-sol} simplifies to $p^*(\tau|\mathbf{w}_r, \mathbf{w}_c) \propto \exp(\mathbf{w}_r^\intercal\Phi_r(\tau) - \lambda\mathbf{w}_c^\intercal\Phi_c(\tau))$ where $\lambda=1+\nu^*\geq 0$. %as $\nu^*\geq 0$.
% \begin{equation*}
%     p^*(\tau|\mathbf{w}_r, \mathbf{w}_c) \propto \exp(\mathbf{w}_r^\intercal\Phi_r(\tau) - (1+\nu^*)\mathbf{w}_c^\intercal\Phi_c(\tau))
% \end{equation*}
% \begin{equation}
% \label{exp-form}
%     p^*(\tau|\mathbf{w}_r, \mathbf{w}_c) \propto \exp(\mathbf{w}_r^\intercal\Phi_r(\tau) - \lambda\mathbf{w}_c^\intercal\Phi_c(\tau))
% \end{equation}

Thus, we observe the probability distribution over a trajectory is the Boltzmann distribution, according to the maximum entropy principle. This supports the intuition that the higher the reward and the lower the constraint value of a trajectory, the more likely it will be. Further, it is seen that $\lambda \geq 0$ acts as a  Lagrange multiplier for  \eqref{opt-pol}. Now, we   define a  partition function $Z(\mathbf{w}_r, \mathbf{w}_c) = \sum_{\text{all}\ \tau} \exp(\mathbf{w}_r^\intercal\Phi_r(\tau) - \lambda\mathbf{w}_c^\intercal\Phi_c(\tau))$ as a  proportionality constant and define:
\begin{equation}
    p^*(\tau|\mathbf{w}_r, \mathbf{w}_c) = \frac{1}{Z(\mathbf{w}_r, \mathbf{w}_c)} \exp(\mathbf{w}_r^\intercal\Phi_r(\tau) - \lambda\mathbf{w}_c^\intercal\Phi_c(\tau)).
\end{equation}

\subsection{Solving the optimization problem}
Clearly,  \eqref{goal-rc} now becomes:
\begin{equation}
\label{mle-form}
\begin{aligned}
    (\mathbf{w}^*_r, \mathbf{w}^*_c) = \text{arg}\max_{\mathbf{w}_r,     \mathbf{w}_c} \prod_{\tau \in D} p^*(\tau|\mathbf{w}_r, \mathbf{w}_c)\quad\quad
    \text{s.t.} \quad \mathbf{w}_c^\intercal \hat{\Phi}_c \leq 1
\end{aligned}
\end{equation}
Here, we once again enforce the constraint that the constraint value function should not exceed the budget so as to not allow arbitrary choices of $\mathbf{w}_r$ and $\mathbf{w}_c$ that would still produce the same $\mathbf{w}_r^\intercal\Phi_r(\tau) - \lambda\mathbf{w}_c^\intercal\Phi_c(\tau)$ as the original $\mathbf{w}_r,\ \mathbf{w}_c$. Simplifying \eqref{mle-form} and taking the log-likelihood, the optimization becomes:
\begin{equation}
    \label{log-like}
    \begin{aligned}
        &\min_{\mathbf{w}_r, \mathbf{w}_c} \underbrace{\left[\frac{1}{|D|}\sum_{\tau \in D} \left(-\mathbf{w}_r^\intercal\Phi_r(\tau) + \lambda\mathbf{w}_c^\intercal\Phi_c(\tau) \right)\right] + \log Z(\mathbf{w}_r, \mathbf{w}_c)}_{\doteq \mathcal{L}} \quad \text{s.t.}\quad \mathbf{w}_c^\intercal \hat{\Phi}_c \leq 1
    \end{aligned}
\end{equation}

The above form in \eqref{log-like} has a convex objective, but the constraint is non-linear and non-convex since $\hat{\Phi}_c$ involves an expectation over the distribution $p^*(\tau|\mathbf{w}_r, \mathbf{w}_c)$. Hence, we use an alternating optimization-based technique where this expectation is computed for a given $(\mathbf{w}_r, \mathbf{w}_c)$ pair, and then \eqref{log-like} is solved as a convex optimization in $(\mathbf{w}_r, \mathbf{w}_c).$  However, evaluating the partition function is computationally difficult as the number of trajectories grows at an exponential rate with $T$. Hence, we use gradient-based approaches. The gradient of the objective function $\mathcal{L}$ in \eqref{log-like} is:
\begin{equation}
\label{gradient}
\begin{aligned}
    &\nabla_{\mathbf{w}_r} \mathcal{L} = -\frac{1}{|D|}\sum_{\tau\in D} \Phi_r(\tau) + \sum_{\text{all}\ \tau}\frac{\exp(\mathbf{w}_r^\intercal\Phi_r(\tau) - \lambda\mathbf{w}_c^\intercal\Phi_c(\tau))}{Z(\mathbf{w}_r,\mathbf{w}_c)}\cdot \Phi_r(\tau)\\
    & \quad\quad = -\Tilde{\Phi}_r + \hat{\Phi}_r\\
    &\text{Similarly, } \nabla_{\mathbf{w}_c} \mathcal{L} = \lambda(\Tilde{\Phi}_c - \hat{\Phi}_c).
\end{aligned}
\end{equation}
Note that, the inverse RL agent can compute $\tilde{\Phi}_r$ and $\tilde{\Phi}_c$ from the demonstrated trajectories, and can compute $\hat{\Phi}_r$ and $\hat{\Phi}_c$ for a given policy.

Now, we can formulate an algorithm to recover both $\mathbf{w}_r$ and $\mathbf{w}_c$. The essential idea is that starting from a random value of $\mathbf{w}_r$ and $\mathbf{w}_c$, we compute the optimal policy corresponding to it. Using the optimal policy, we calculate the state visitation frequency and hence, the PFE. After that, we apply a gradient step on $\mathbf{w}_r$ and $\mathbf{w}_c$. We alternate between the two steps till the change in $\mathbf{w}_r$ and $\mathbf{w}_c$ between two successive iterations are within a desired tolerance level. Algorithm \ref{algo:full-problem} provides an outline of the proposed scheme. In steps $10$ and $11$ of the algorithm, we project $\mathbf{w}_r$ and $\mathbf{w}_c$ to a normalized set of $1$-norm unity with non-negative coordinates using the usual Bregman divergence minimization~\cite{hazan-oco} with negative entropy function $f(\mathbf{x})=\sum_i \mathbf{x}_i \log(\mathbf{x}_i)$ for numerical stability.
Also, note that the use of Exponentiated Gradient Descent as the intermediate gradient step in \eqref{log-like} requires such a projection for regularization (see sec. 5 of~\cite{hazan-oco} for details). However, other forms of gradient updates can also be used with different regularizations. Once the algorithm computes optimal $\mathbf{w}_r$ and $\mathbf{w}_c$, the inverse RL agent can recover $r(s)=\mathbf{w}_r^\intercal \Phi_r(s),\ c(s)=\mathbf{w}_c^\intercal \Phi_c(s)$.

%This has been illustrated in fig.\ref{fig:alternate-max-grad}.

% \begin{figure}[t]
% \centering
% \begin{tikzpicture}[node distance = 20mm, on grid, auto]
% \node (step1) [startstop] {From current estimate of $\mathbf{w}_r$ and $\mathbf{w}_c$, recover the (approx.) optimal policy $\pi$ and the optimal Lagrange multiplier $\lambda$. Then calculate $d_t(s)$ for all states.};
% \node (step2) [startstop, below of = step1] {With current estimate of $d_t(s)$, solve optimization problem \eqref{log-like} to get new values of $\mathbf{w}_r$ and $\mathbf{w}_c$ (using gradient based methods)};

% \draw [arrow] (step1.west) to [out = 225, in=135] (step2.west);
% \draw [arrow] (step2.east) to [out = 45, in=315] (step1.east);
% \end{tikzpicture}
% \caption{Scheme for recovery of the reward and constraint}
% \label{fig:alternate-max-grad}
% \end{figure}

\begin{algorithm}[t]
%\setstretch{1.7}
\caption{Iterative Algorithm for Reward and Constraint Recovery}
\label{algo:full-problem}
\begin{algorithmic}[1]
\Require $p_0,\ p(s'|s, a),\ D,\ \text{learning rate}=\kappa>0$
\State Initialize $\mathbf{w}_r \in \mathbb{R}^{d_r}$ and $\mathbf{w}_c \in \mathbb{R}^{d_c}$ such that $\lvert\mathbf{w}_r\rvert_2 = \lvert\mathbf{w}_c\rvert_2 = 1$
\State Compute $\Tilde{\Phi}_r$ and $\Tilde{\Phi}_c$ using eq.\eqref{efe-pfe}
\Repeat
\State $\pi, \lambda \gets$ Optimal Policy and Optimal Lagrange Multiplier for CMDP$(\mathbf{w}_r, \mathbf{w}_c)$
\State $\text{Calculate}\ d_t(s)\ \forall\ t=1\dots, T,\ \forall\ s\in S\ \text{under}\ \pi\ \text{using eq.\eqref{svf-recur}}$
\State Compute $\hat{\Phi}_r$ and $\hat{\Phi}_c$ using eq.\eqref{pfe-svf}
\State Compute the gradients $\nabla_{\mathbf{w}_r}\mathcal{L}$ and $\nabla_{\mathbf{w}_c}\mathcal{L}$ using eq.\eqref{gradient}
\State $\mathbf{w}_r \gets \mathbf{w}_r \cdot \exp(-\kappa \nabla_{\mathbf{w}_r}\mathcal{L})$
\State $\mathbf{w}_c \gets \mathbf{w}_c \cdot \exp(-\kappa \nabla_{\mathbf{w}_c}\mathcal{L})$
\State $\mathbf{w}_r \gets \text{arg}\min_\mathbf{w}\ \sum_{i}[\mathbf{w}]_i\log([\mathbf{w}]_i/[\mathbf{w}_r]_i)\ \text{s.t}\ \sum_i [\mathbf{w}]_i=1, \mathbf{w} \geq 0$
\State $\mathbf{w}_c \gets \text{arg}\min_\mathbf{w}\ \sum_{i}[\mathbf{w}]_i\log([\mathbf{w}]_i/[\mathbf{w}_c]_i)\ \text{s.t}\ \sum_i [\mathbf{w}]_i=1,\ \mathbf{w} \geq 0,\ \mathbf{w}^\intercal\hat{\Phi}_c \leq 1$
\Until {Convergence}
\State \textbf{Return} $\mathbf{w}_r\,\ \mathbf{w}_c,\ \pi$
\end{algorithmic}
\end{algorithm}

\section{Experiments}
In this section, we use the gridworld environment  to evaluate the proposed algorithm. We consider the grid of size $5\times5$. The agent always starts from the top left corner of the grid. At any time step, the agent has four actions available: up, down, left, and right. An action has 70\% probability of success while with 30\% probability, the agent moves into any of the neighbouring blocks. This adds stochasticity to the transition, which makes learning from data more difficult. The states feature vectors are: $\Phi_r((x,y)) = (2.5\times 10^{-3})\begin{bmatrix} x, & y \end{bmatrix}^\intercal$ and $\Phi_c((x,y)) = 0.1 \begin{bmatrix} e^{-c_1 (h_1 - x)}, & e^{-c_2 (h_2 - y)}, & \min\{\lvert 4-x\rvert, x\}, & \min\{\lvert 4-y\rvert, y\}\end{bmatrix}^\intercal$. The need for powerful representation for MaxEnt IRL\cite{maxentirl}  is well-known.

\begin{figure}[t]
\centering
\begin{subfigure}{0.35\textwidth}
    \centering
    \includegraphics[width=\textwidth]{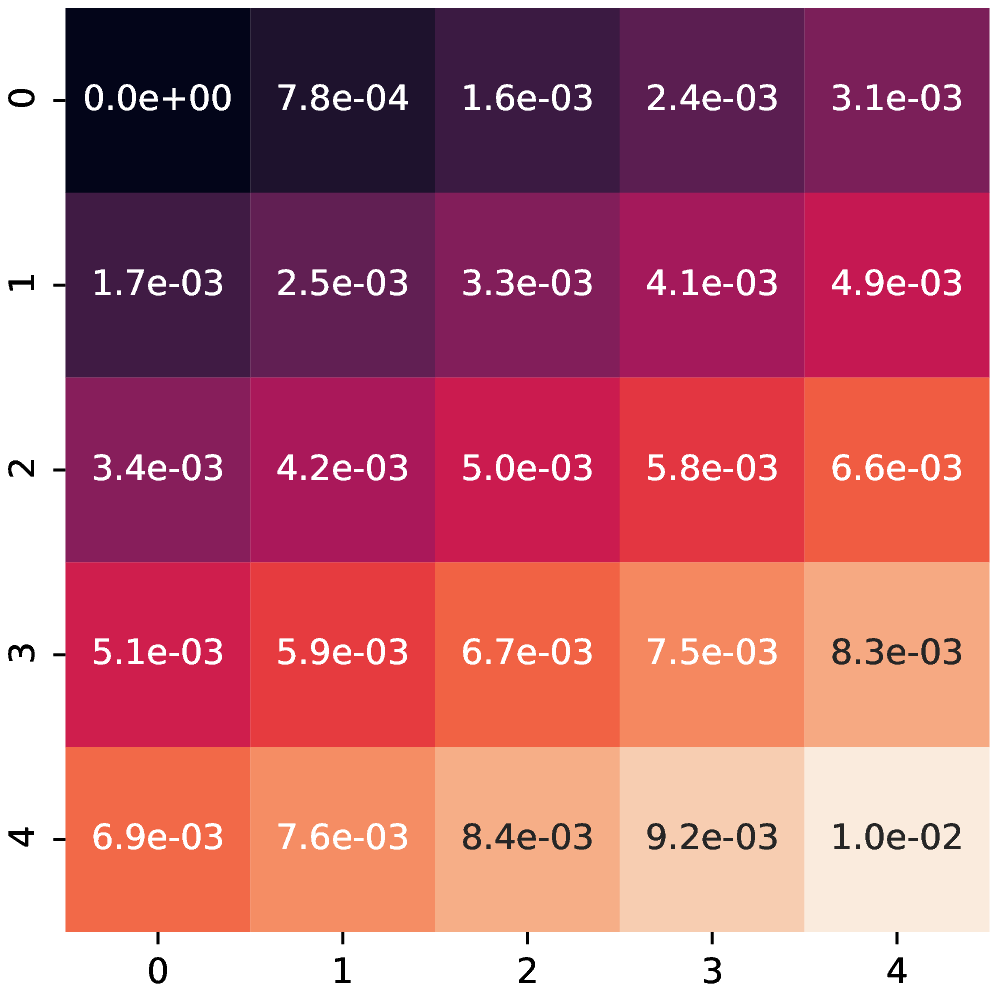}
    \caption{True Reward}
\end{subfigure}
\hfill
\begin{subfigure}{0.35\textwidth}
    \centering
    \includegraphics[width=\textwidth]{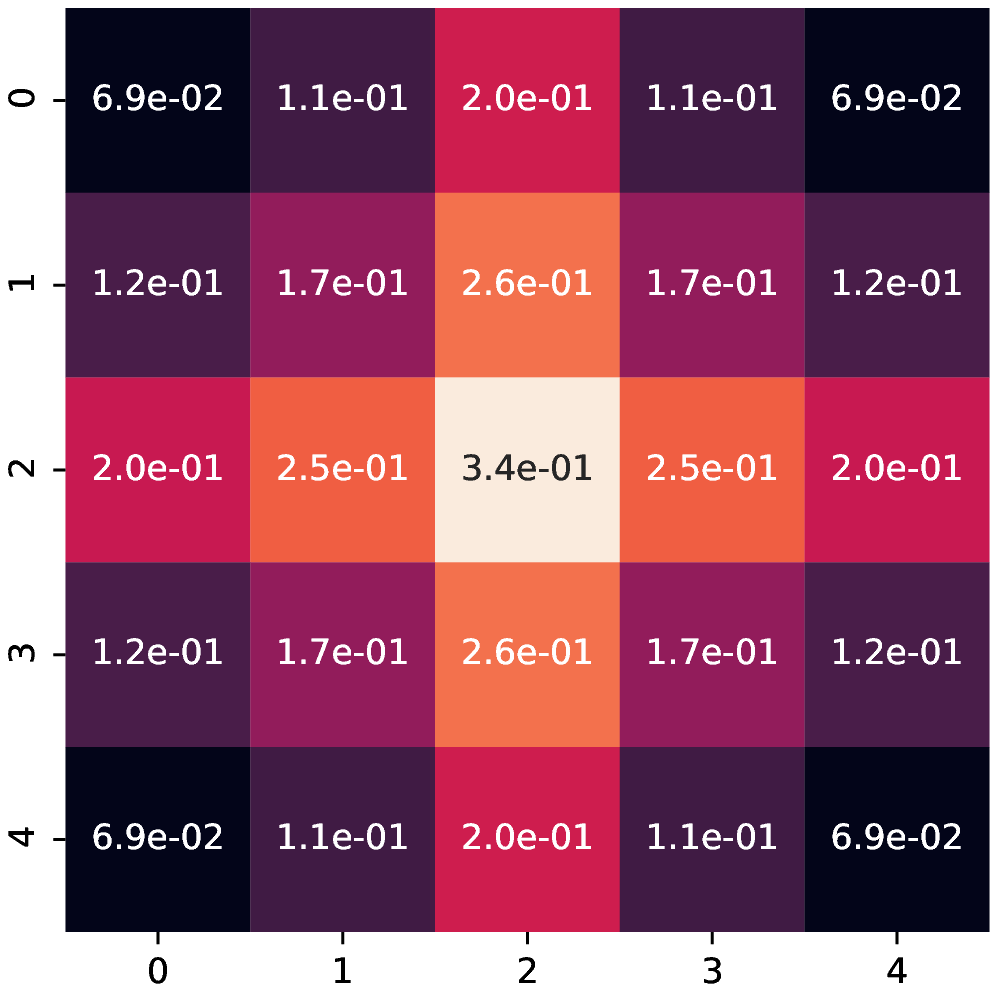}
    \caption{True Constraint}
\end{subfigure}
\hfill
\begin{subfigure}{0.28\textwidth}
    \centering
    \includegraphics[width=0.8\textwidth]{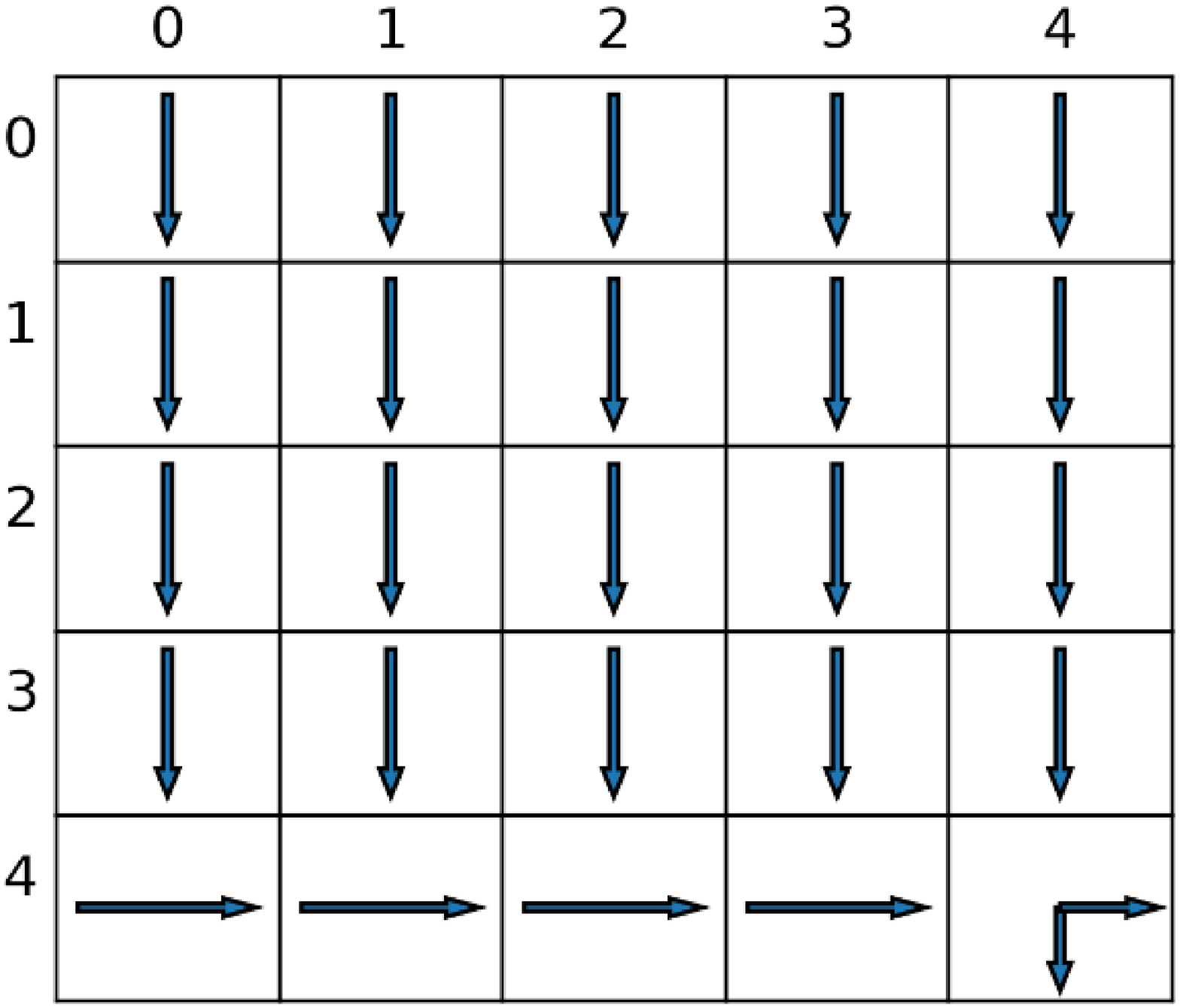}
    \caption{True Optimal Policy}
\end{subfigure}
\\
\begin{subfigure}{0.35\textwidth}
    \centering
    \includegraphics[width=\textwidth]{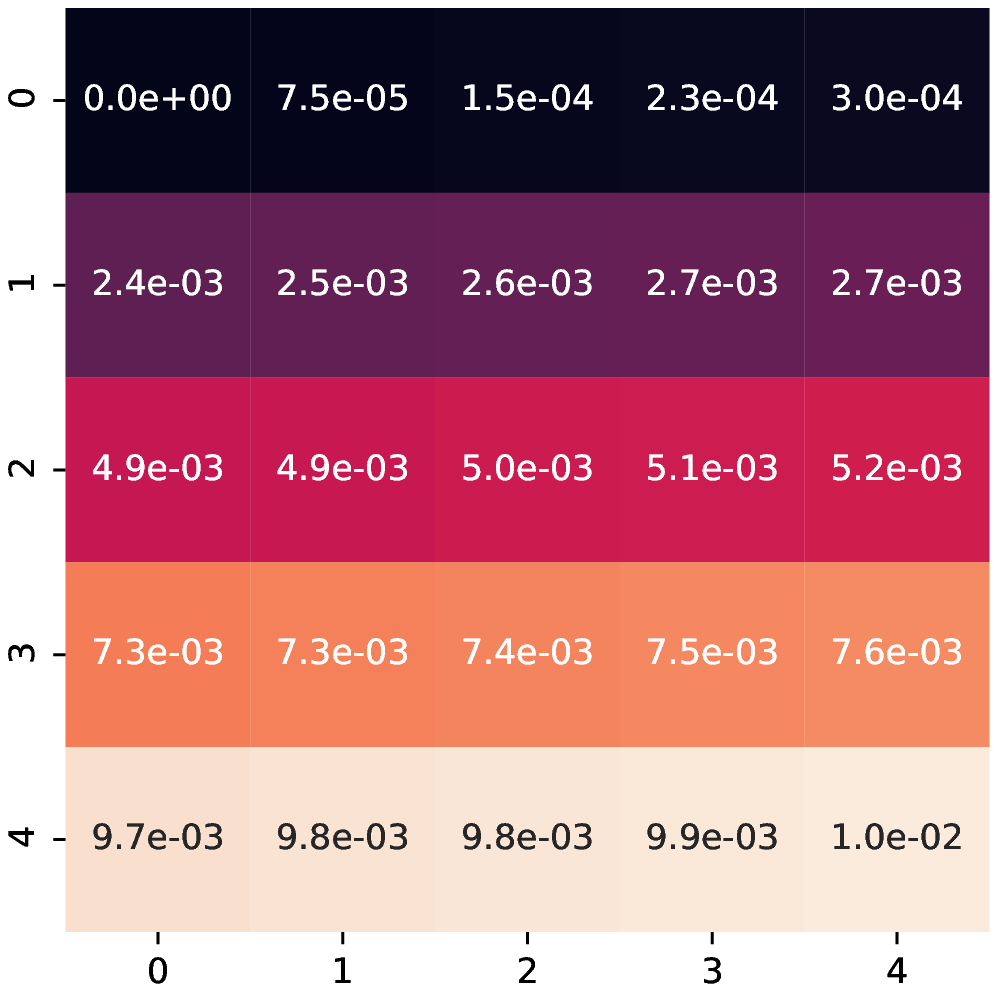}
    \caption{Predicted Reward}
\end{subfigure}
\hfill
\begin{subfigure}{0.35\textwidth}
    \centering
    \includegraphics[width=\textwidth]{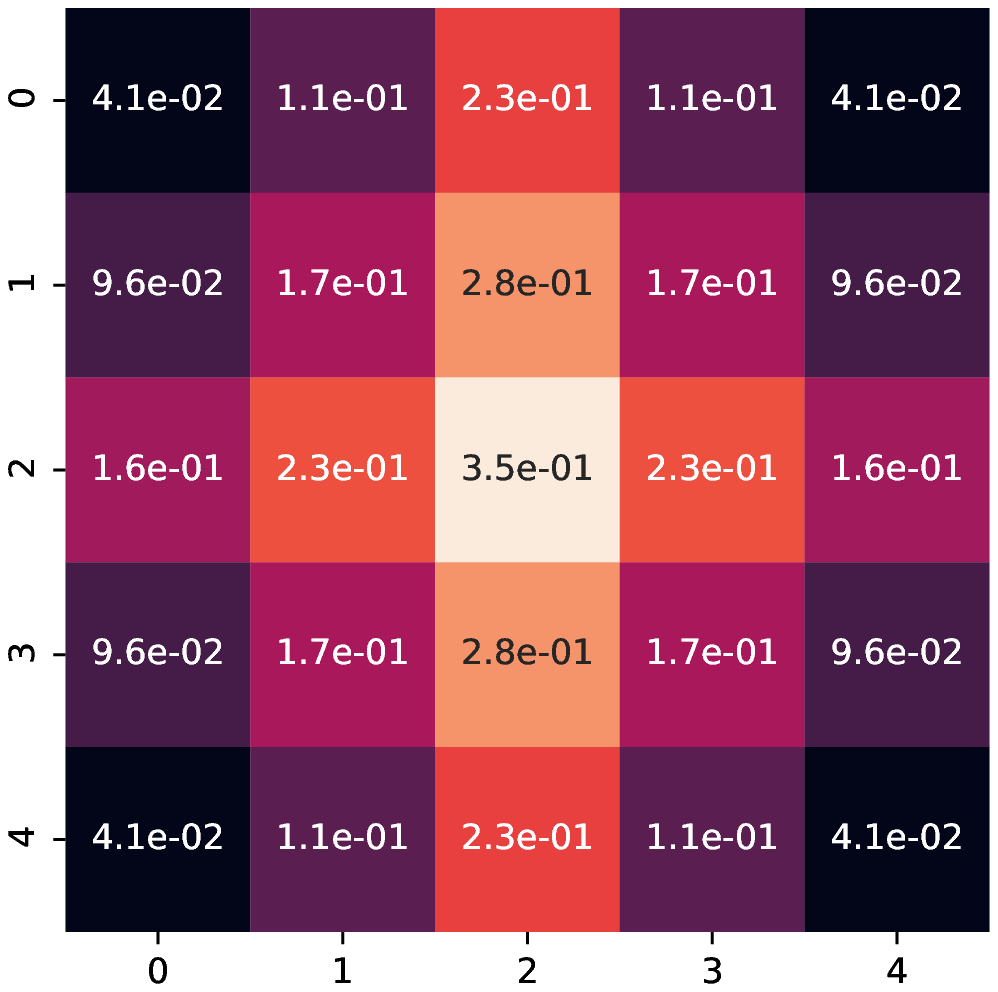}
    \caption{Predicted Constraint}
\end{subfigure}
\hfill
\begin{subfigure}{0.28\textwidth}
    \centering
    \includegraphics[width=0.8\textwidth]{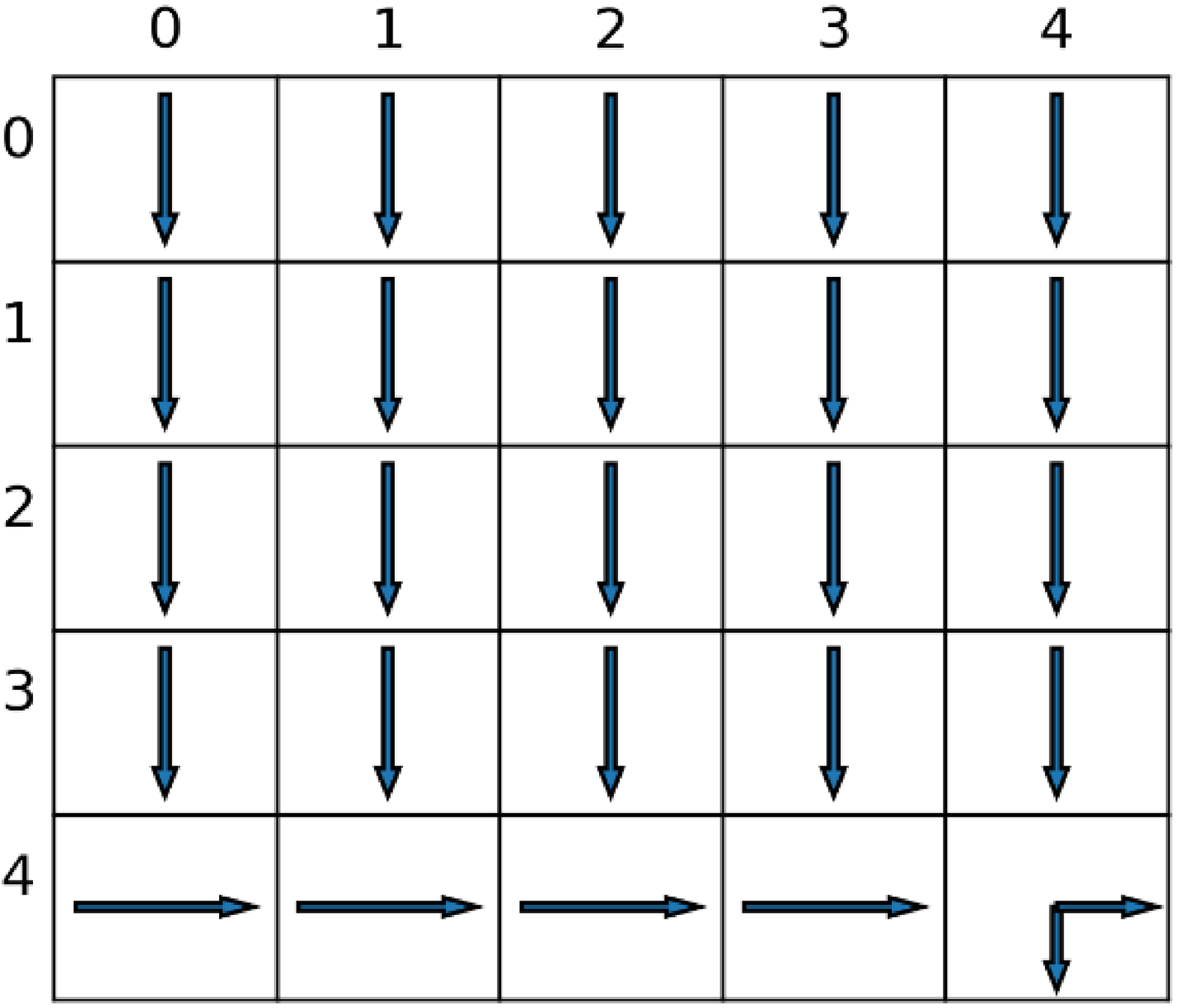}
    \caption{Predicted Policy}
\end{subfigure}
    \caption{Pictorial demonstration of the performance of our algorithm in a grid-world setting. The numeric values of rewards and costs are written inside the boxes. The arrows denote the optimal action in a state. }
    \label{fig:result}
\end{figure}

We choose $(c_1, c_2) \sim uniform[0,1]^2$ uniformly at random from the unit square,  and $(h_1, h_2)$ is chosen uniformly at random from the set of tuples $\{(i,j): 1 \leq i, j \leq 3,\ i,j\in \mathbb{Z}\}$. The variable $(h_1, h_2)$ corresponds to the location of a ``hill" with ``slopes" $(c_1, c_2)$. In Figure~\ref{fig:result}, we show results only for one realization of $(c_1,c_2,h_1,h_2)$, though similar patterns were observed for numerical experiments using many other realizations.  
 The true reward and the constraint parameters are chosen to be $\mathbf{w}_r = [q, 1-q]^\intercal,\ q\sim uniform[0,1]$ and $\mathbf{w}_c = [a, 1-a, b, 1-b]^\intercal,\ a, b \sim uniform[0,1]$. This has been done to keep the reward and constraint values normalized. We use $100$ samples with $T=2000$.  

In Figure~\ref{fig:result}, we can see that the reward and the constraints have been recovered well qualitatively, and the optimal policy for the recovered rewards and constraint is the same as the true policy. The simulation was run for $10$ seeds, and the same trends were observed in each of them. This shows the effectiveness of our method, which is able to recover both the reward and the constraint from the given demonstrations. While experimenting, we observed that increasing $T$ increased the accuracy of our method. This happens  because the policy used to generate the trajectory corresponds to a stationary policy optimal for the infinite horizon discounted cost CMDP. Obviously, the longer the trajectory, the better it describes the policy. We also notice that the recovered reward follows the same pattern as the original reward (increasing from left to right and from top to bottom). However, the variation in numeric values in the recovered reward function is smaller, possibly because the principle of maximum entropy and normalization of $\mathbf{w}_r$ prevent any additional bias than what is strictly necessary to obtain the same data distribution as in the data set. Furthermore, we see that the peak of the penalty cost function (lightest colour) is well recovered through our method. Hence, we qualitatively demonstrate the effectiveness of our method. Since no previous work deals with the case of recovering both the reward and the constraint, we could not provide any comparison with other works.

\section{Conclusion and Future Work}
In this work, we focused on IRL-CR in the offline setting, in which the trajectory data is available in a batch. Though the problem is a difficult one and ill-posed, we have proposed a low-complexity algorithm to solve it.  There has been no prior work on IRL-CR. Hence, there is a lack of a previous baseline to compare the performance of our work. However, we demonstrate strong results on a qualitative basis.  

One immediate future step is to extend the framework to the settings where the data arrives in an online fashion.  Also, though our work assumes known features for the states, this information may not be available in a real-world setting, and representation learning needs to be employed along with IRL-CR. Also, providing provable performance guarantees to the proposed algorithms will be very challenging, given that the problem is ill-posed. We will address these problems in our future research endeavours.

\bibliographystyle{splncs04}
\bibliography{references}
\end{document}